\documentclass[letterpaper, 10 pt, conference]{ieeeconf}  

\IEEEoverridecommandlockouts                              

\usepackage[pdftex]{graphicx}
\usepackage{amsmath} 
\usepackage{amssymb}  
\usepackage{xcolor}

\usepackage[normalem]{ulem}
\usepackage[rightcaption]{sidecap}

\usepackage{fancyhdr}

\fancypagestyle{IEEEcopyright}{%
  \fancyhf{}
  
  \fancyfoot[L]{\fontsize{8}{10}{\sffamily{\copyright 2020 IEEE. Personal use of this material is permitted. Permission from IEEE must be obtained for all other uses, in any current or future media, including reprinting/republishing this material for advertising or promotional purposes,creating new collective works, for resale or redistribution to servers or lists, or reuse of any copyrighted component of this work in other works.}
  }
}
}
\newcommand{\eg}{\textit{e.g.,~}}
\newcommand{\ie}{\textit{i.e.,~}}

\newcommand{\fref}[1]{Fig \ref{fig:#1}}

\newcommand{\add}[1]{#1}

\title{\LARGE \bf Evolving Dyadic Strategies for a Cooperative Physical Task}

\author{Saber Sheybani$^{1\dagger}$, Eduardo J. Izquierdo$^{2}$, and Eatai Roth$^{1}$
\thanks{$^{1}$Department of Intelligent Systems Engineering, Indiana University, Bloomington, IN 47406}%
\thanks{$^{2}$Cognitive Science Program, Indiana University, Bloomington, IN 47406}%
\thanks{$^{\dagger}$ corresponding author, sheybani@iu.edu}%
}

\begin{document}


\maketitle


\thispagestyle{IEEEcopyright}

\begin{abstract}
Many cooperative physical tasks require that individuals play specialized roles (\eg leader-follower). Humans are adept cooperators, negotiating these roles and transitions between roles innately. Yet how roles are delegated and reassigned is not well understood. Using a genetic algorithm, we evolve simulated agents to explore a space of feasible role-switching policies. Applying these switching policies in a cooperative manual task, agents process visual and haptic cues to decide when to switch roles. We then analyze the evolved virtual population for attributes typically associated with cooperation: load sharing and temporal coordination. We find that the best performing dyads exhibit high temporal coordination (anti-synchrony). And in turn, anti-synchrony is correlated to symmetry between the parameters of the cooperative agents. These simulations furnish hypotheses as to how human cooperators might mediate roles in dyadic tasks.\\

\end{abstract}


\section{INTRODUCTION} Physical interactions between people are commonplace: two colleagues shake hands, a physical therapist guides her patient's arm to train a new movement skill, or a delivery person hands a package to their customer. Some interactions demand similar behaviors from the participants (\eg the handshake). In other instances, cooperation requires individuals to assume distinctly different roles (\eg the leader-follower relationship of the therapist and patient). And roles may change over the course of a task (\eg the hand-off between the delivery person and recipient). Dyadic interactions require coordination between individuals: an agreement on the task objective, the delegation of roles, and possibly, the re-assignment of these roles. For humans, this subliminal dialogue is innate.

Sebanz et al. proposed that the mechanisms that underlie joint action cannot be inferred solely from the controllers of the constituent participants \cite{sebanz_joint_2006}, that the interaction is categorically more than the sum of its parts. Role specialization is one such emergent feature of cooperative tasks. Reed et al. paired human subjects in a physical joint action task in which subjects were instructed to turn a wheel to a prescribed angle by pushing or pulling on opposite ends of a crank \cite{reed_physical_2008, reed_haptically_2006}. In this paradigm, dyads exhibited role specialization and better performance than individuals performing the same task alone; despite this, subjects largely reported that their partner was a hindrance. Reed proposed a haptic channel of communication---participants may rely on haptic cues (either through direct contact or as force transmitted through a mutually manipulated object) to solve role delegation. In cooperative tasks, dyads exert overlapping (antagonistic forces) that don't contribute to achieving the task goal \cite{van_der_wel_let_2011}; is this an inefficiency of coordination or evidence of haptic communication? 

Stefanov et al. proposed a similar paradigm, conductor-executor, suggesting that one agent assumes responsibility for the planning and decision making and the other is largely responsible for enacting the plan \cite{stefanov_role_2009}. And Groten et al. observed that human dyads prefer some dominance difference (unequal control) \cite{groten_experimental_2009}. Though Madan et al. also advocate this haptic communication channel, they observe a spectrum of interactions (from harmonious to conflicting) shaped by the task and the states of the participants \cite{madan_recognition_2015}. So, while there are numerous hypotheses as to what roles cooperators assume, there is good evidence that haptic cues contribute to how they are delegated.    

\add{This work aims at understanding how individuals decide which role to assume (and when to switch role) as a function of visual and haptic sensory cues. Our goal is not to design an optimal strategy for role delegation, but rather to discover a variety of feasible strategies \cite{browning2004evolution, panait2005cooperative}, as one might expect to find in a sample population of human subjects.}

\add{As a means of exploring the parameter space of role-switching policies, we apply a multi-objective evolutionary algorithm to dynamical simulations of a cooperative manual task.} 
\add{In simulation, virtual dyads (pairs of agents) cooperate to push an egg (a fragile object) along a prescribed trajectory (\fref{traces*}a). But dyads must also be careful to regulate the applied forces so as not to drop or crush the egg. These force constraints give rise to two distinct roles that we refer to as reference-tracking (T) and force-stabilization (S); to perform the task successfully, each agent comprising a dyad must apply rules based on visual and haptic sensory cues to decide when to perform each role.}

\add{The multi-objective evolution furnishes a varied set of solutions along the Pareto frontier, revealing common qualities among high performing cooperative dyads. We might intuitively ascribe some attributes of dyadic behavior as indicators of cooperation: \emph{load sharing}---agents contribute effort equitably---and \emph{anti-synchrony}---agents assume complementary roles with temporally coordinated role swapping. An analysis of the Pareto-efficient sub-population reveals that the best performing dyads exhibit greater anti-synchrony. In contrast, load-sharing emerged ubiquitously, not correlated to performance. We further conjecture that \emph{symmetry} (parameter-level similarity between paired agents' switching policies) is a genomic pattern that underlies cooperation.}

\add{Further exploration of the parameter space may identify patterns of cooperative dyads; can evolved dyads predict archetypes of cooperative strategies in humans? Although these simulations are not matched to a specific physical experiment, this approach serves as a framework for generating hypotheses in preparation for human subject experiments.}

\begin{figure*}[tbh]
 \centering
 \includegraphics[width=\textwidth]{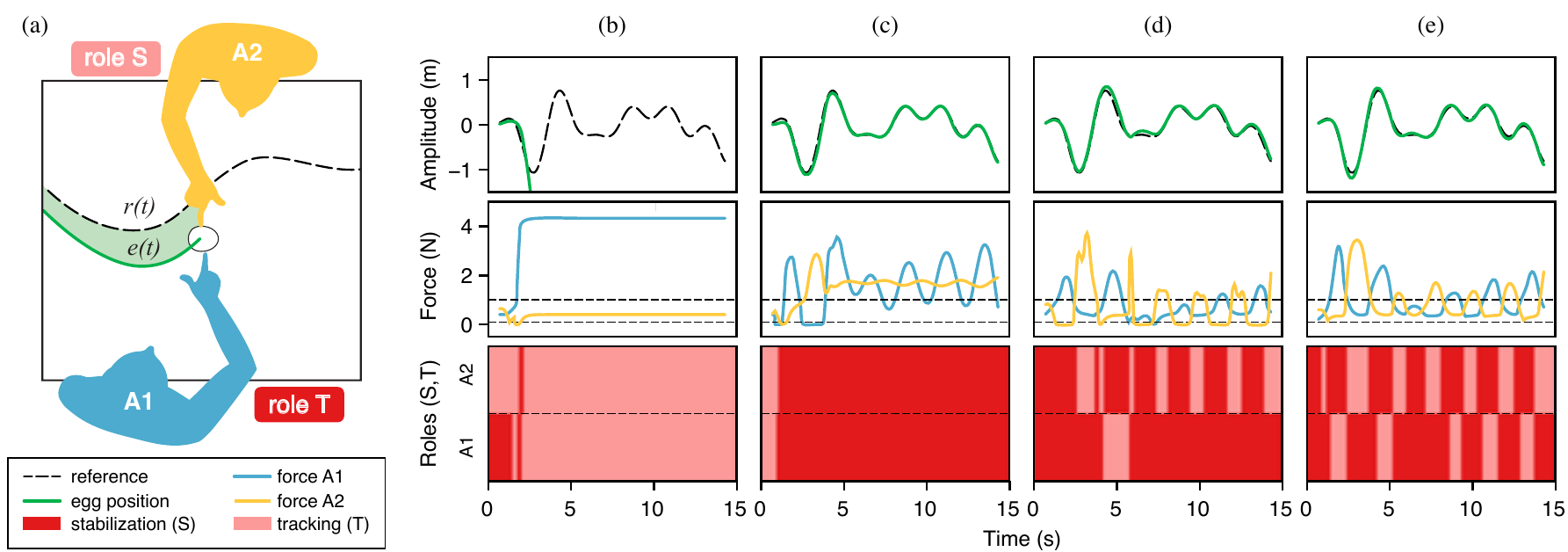}
 \caption{(a) Agents coordinate opposing forces to move the egg along the prescribed trajectory $r(t)$. (b-e) \emph{(Top)} 
 The trajectory of the egg (\emph{green}) and the target trajectory (\emph{dashed black}), \emph{(Middle)} the force contributions of each agent (\emph{blue} and \emph{gold}) and the force constraints (\emph{dotted black}), and \emph{(Bottom)} the roles of each agent. (b) A dyad fails to track the trajectory, (c) a dyad succeeds at tracking but violates the force constraint, (d) a dyad exhibits some role switching and infrequently violates the force constraints, and (e) a dyad performs the task rather successfully.}
 \label{fig:traces*}
\end{figure*}

\section{METHODS}

We explore a space of linear role-switching policies using a multi-objective evolutionary algorithm implemented using the DEAP package \cite{fortin_deap:_2012} in Python. In each generation (iteration of the evolutionary algorithm), dyads perform a joint action tracking task and were evaluated on tracking performance, force regulation (whether the egg remains intact), individual energy expenditure, and smoothness of the force profile.

\subsection{Task Design} \label{sec:task-structure}

In the simulated joint tracking task, a pair of agents (the dyad) guide an ``egg" along a prescribed trajectory, $r(t)$ (\fref{traces*}). The agents push in opposing directions, modulating their individual forces, $f_1$ and $f_2$, such that the net force moves the egg (similar to the executor-conductor task in \cite{stefanov_role_2009}). Additionally, the agents must regulate the normal force on the egg, $f_N$: too low and the egg is dropped, too high and it is crushed. The task dynamics and force constraints are:
\begin{gather}
    m \ddot{x} + b \dot{x} = f_1 - f_2\\
    f_{min} < f_N = \min (f_1, f_2) < f_{max}
\end{gather}

\noindent where $m = 0.5 ~kg$ and $b = 1~N \cdot s \cdot m^{-1}$ are the egg's mass and damping coefficient, respectively, and $f_{min} = 0.1~N$ and $f_{max} = 1~N$ are the bounds on applied force.

These force constraints dictate two distinct roles: the less forceful agent is solely responsible for regulating the normal force within the allowable range, whereas the agent that pushes harder largely controls the trajectory of the egg. We will refer to these two roles as normal force \emph{stabilization} (S) and reference \emph{tracking} (T). The reference trajectories are designed such that dyads cannot perform the task successfully without switching roles. \add{If the reference can be tracked with both agents' forces within the allowable force limits, both agents can assume the tracking role (T) and avoid switching. Hence, for trajectories that permit the non-switching strategies, the net force can be bound by:}
\begin{gather}
    f_{min} < f_1, f_2 < f_{max} \qquad \text{(bounds on individuals)}\\
    |f_1-f_2| < f_{max}-f_{min} \qquad \text{(bounds on net force)}\label{eq:forcebound}
\end{gather}

\add{For a given reference trajectory, suppose $f_{ideal}$ is the net force signal, $f_1-f_2$, that tracks the trajectory perfectly; we design trajectories such that this ideal force violates the constraint (Eq \ref{eq:forcebound}) frequently.}


For a given trial, the reference trajectory is selected from a set of pre-designed sum-of-sines signals (five sinusoidal components between $0.1-0.5~Hz$) normalized to require the same amount of control effort. Because each agent only applies force in one direction (pulling is disallowed by the force constraint), dyads should swap roles roughly whenever the acceleration of the reference trajectory changes sign; this observation suggests a candidate policy. However, the task permits solutions with diverse switching policies and timing.

\subsection{Anatomy of a dyad}

In our evolutionary algorithm, members of the population represent dyads, each with two constituent agents who co-evolve. Each agent possesses two control policies---one for reference tracking and another for force regulation---and two role-transition rules that govern when the agent should switch from using one control policy to the other---S-to-T and T-to-S transitions.

Both the control policies, $C$, and role-transition rules, $W$, are linear functions of a feature vector, $\theta \in \mathbb{R}_{9\times1}$:
\begin{equation}
    \Theta = \begin{bmatrix}
        r & r' & r'' & e & e' & e'' & f_N & f'_N & 1
    \end{bmatrix}
\end{equation}

Such that:
\begin{align}
    C\cdot\Theta &= \dot{f} \quad &\text{(control policy)}\\ 
    W\cdot\Theta &> 0 \quad &\text{(role-transition rule)}
\end{align}
\noindent comprising observations of the reference trajectory (and its first and second derivatives), the tracking error (and its first and second derivatives), the normal force (and its first derivative), and a bias term. These features mimic the visual and haptic cues humans might attend to in a cooperative manual task.

Each agent is randomly assigned controllers from a set of pre-designed policies; reference-tracking controllers are PID and force-regulation controllers are integral-only. The control policies govern how the agents should modulate their force output. The output of the controller ($C\in\mathbb{R}_{9x1}$) is $\dot{f}$ to effect smooth force trajectories.

Force-stabilization controllers have the form:
\begin{equation}
\begin{gathered}
     K_I(f_N - f_{opt}) = \dot{f} \longrightarrow \\
     C_S=
     \begin{bmatrix}
        0&0&0&0&0&0&-K_I&0& K_I f_{opt}
    \end{bmatrix}
 \end{gathered}
 \end{equation}
 
\noindent and reference-tracking controllers have the form:
\begin{equation}
\begin{gathered}
    K_I\cdot e + K_P \cdot \dot{e} + K_D \cdot  \ddot{e} = \dot{f} \longrightarrow\\
    C_{T}=
    \begin{bmatrix}
        0&0&0&K_I&K_P&K_D&0& 0& 0 
    \end{bmatrix}\\
\end{gathered}
\end{equation}

\noindent The controllers are \emph{not evolved} in our algorithm, but each designed controller is sufficient for performing its task (either S or T). These agents can perform their roles individually, but dyads must discover rules for cooperation.

These rules are encoded as in the role-transition policies, encoded in the weight vector $W \in \mathbb{R}_{9 \times 1}$:
\begin{equation}
        W_{ST}\dot\Theta > 0 \quad \text{and} \quad W_{TS}\dot\Theta > 0
\end{equation}

Because inequalities are invariant to positive scaling, switching policies are normalized so that $|W|_2 = 1$; equivalent policies are all mapped to the same point on the sphere $S^9$. \emph{Our algorithm evolves these role-transition policies.} 

The dyad is represented as an array comprising the switching policies and controllers of each agent:
\begin{equation}
    P = \begin{bmatrix}
        W_{1,ST} & W_{1,TS} & C_{1,S} & C_{1,T}\\
        W_{2,ST} & W_{2,TS} & C_{2,S} & C_{2,T}
    \end{bmatrix}
    \quad
    \begin{matrix}
    \text{(Agent 1)}\\
    \text{(Agent 2)}
    \end{matrix}
\end{equation}

\subsection{Task Simulation}

Each iteration of the dynamical simulation proceeds with the following steps:

\begin{itemize}
    \item Each agent evaluates the transition policy corresponding to their current role (\eg if Agent 1 is in Role S, they will evaluate $W_{ST} \cdot \Theta > 0$). If the switching policy evaluates as True, the agent switches role. \add{To preclude rapid oscillation between roles, each transition is followed by a 250 ms refractory period.}
    \item  Agents apply their role-specific controller to calculate a force output. Force output is restricted to be non-negative (pulling is disallowed).
    \item The net force, $f_1-f_2$, is provided as input to a dynamical simulation (Runge-Kutta fourth-order) to calculate the egg motion.
    \item The normal force, $\min(f_1,f_2)$, is checked to be within bounds.
    \item A new feature vector is constructed. 
\end{itemize}

Note, the agents receive different feature vectors, because they perceive the trajectory and tracking error from opposite directions; they sense the same normal force however. 

\subsection{Objectives}
The evolutionary algorithm evaluates fitness with respect to the following objectives:
\begin{itemize}
    \item \emph{Tracking loss} - the RMS tracking error over a trial.
    \item \emph{Force stabilization} - penalties incurred for exceeding the force bounds on the egg. The stabilization loss is evaluated as a piece-wise linear function, zero within the range of allowable forces, and increasing with constant slope beyond the force bounds.
    \item \emph{Individual Effort} - the RMS force of each agent. These are separate objectives, to preserve genes from both altruistic and selfish agents.
    \item \emph{Jerk} - to preserve genes that generate smooth force outputs. Human motion minimizes jerk \cite{flash_coordination_1985}.
\end{itemize}

\subsection{Culling and Repopulation}
For an initial population, we generated 500 dyads. Each agent in the dyad was randomly assigned stabilizing and reference-tracking controllers. Switching policies are initialized to be sparse; one feature weight is selected randomly and assigned a value of either $1$ or $-1$ and the bias term is randomly selected from the set $\{-0.3, 0.3\}$ (then normalized), resulting in a policy dependent on a single feature.

These initial 500 dyads then produce 500 offspring dyads, either by cross-over or mutation. Cross-over creates an offspring with role-switching policies set as the average (normalized) of the parent dyad policies and control policies are inherited from one of the parent dyads selected randomly. Mutations occur at a rate of 0.05, and generate new dyads \emph{de novo}. Despite initializing with sparse policies, the population evolves more complex policies through cross-over and mutation.

In each iteration, the 1000 dyads are tested on one of a set of reference trajectories. They are scored on the five objectives above and culled back down to 500 using Non-dominated Sorting Genetic Algorithm II (NSGA-II \cite{deb_fast_2002}). Throughout the evolution, Pareto efficient dyads were collected in a ``Hall of Fame".

\begin{figure*}[tbh]
 \centering
 \includegraphics{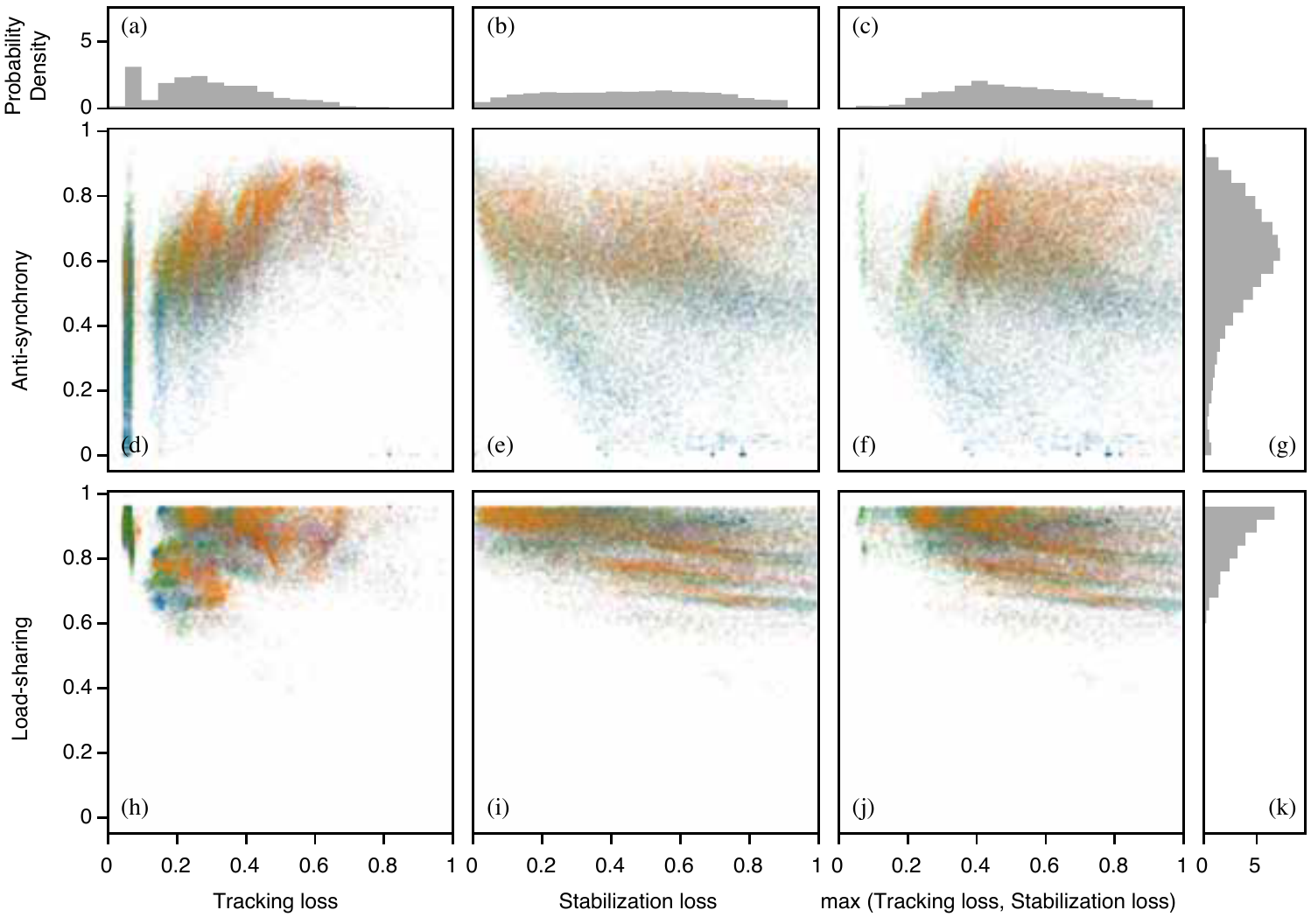}
 \caption{Relating anti-synchrony and load-sharing properties to performance reveals the importance of anti-synchrony to stabilization and the ubiquity of load-sharing in the Pareto-efficient
 population. (d-f) anti-synchrony vs loss functions, (h-j) load-sharing vs loss functions, and histograms of each loss function (a-c) and for anti-symmetry (g) and load-sharing (k) properties. Data points are colored only to denote corresponding groups across the subfigures.}
 \label{fig:pareto-scatter}
\end{figure*}

\section{RESULTS}

\subsection{Evolved role switching}

We are able to evolve many parametrically distinct dyads that accomplish the main goals of the task (\ie tracking the reference without dropping or crushing the egg). Most successful dyads exhibit role-switching to various degrees. Figure \ref{fig:traces*}b-e presents four representative dyads that emerged throughout evolution to demonstrate different modes of failure. For comparison, we simulate these dyads on a new reference trajectory (not a trajectory any had been trained on). Each column corresponds to a dyad with the rows (from top to bottom) illustrating the reference and egg trajectories, the force contributions of each agent, and the agents' respective roles.
In an early evolved dyad (\fref{traces*}b), agents both settle at the stabilization role (S) after a few seconds and consequently fail to track the reference; they do manage to maintain the normal force within the permissible range. The dyad shown in \fref{traces*}c exhibits the opposite, both agents get stuck in the tracking role (T), tracking the reference accurately but consistently violating the force constraints. The dyad in \fref{traces*}d exhibits some role-switching and good tracking performance but intermittently drops below the lower force bound (dropping the egg). The last dyad (\fref{traces*}d) exhibits temporally coordinated role-switching and succeeds at the task.

\begin{figure*}[bth]
 \centering
 \includegraphics{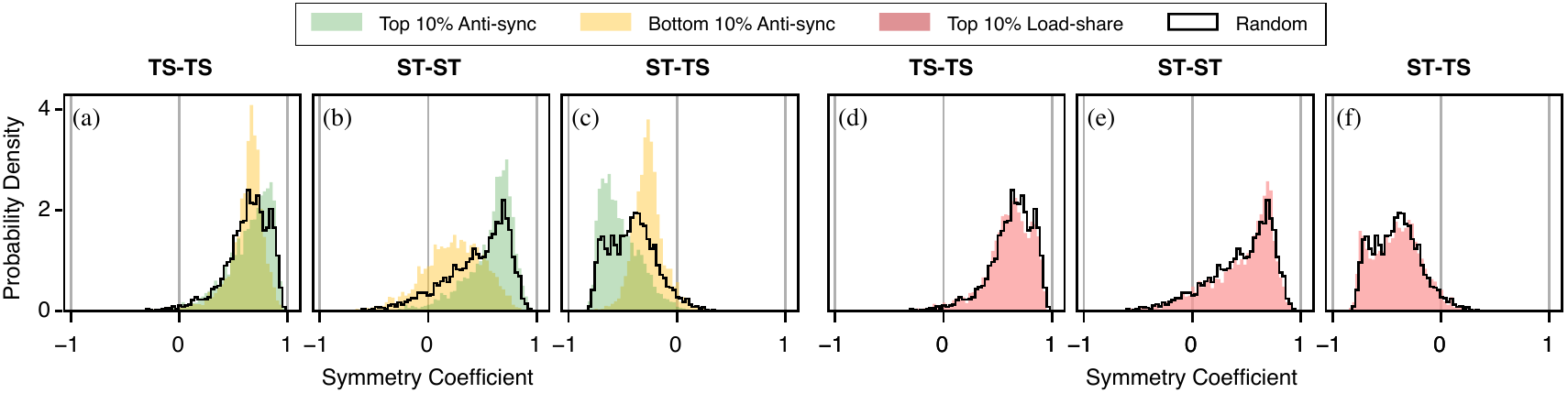}
 \caption{Symmetry of switching policies between agents is pronounced in dyads exhibiting the highest anti-synchrony (a-c, \emph{green}) and less so in the bottom decile (\emph{yellow}). Dyads in the top decile for load-sharing (d-f, \emph{red}) are no more symmetric than a random sampling of the population (\emph{black outline}). }
 \label{fig:symm*}
\end{figure*}


\subsection{Evolved anti-synchrony and load-sharing}

We generate five randomly initialized populations, each comprising 500 dyads evolved over 100 generations \add{(\ie 250000 distinct dyads were tested overall)}. We collect all the distinct Pareto-efficient dyads, filtering out dyads that had tracking loss greater than 1 (they did worse than had they not done anything at all) and stabilization loss of greater than 1 (corresponding to breaking the egg at every time step). 32772 dyads constitute the Pareto-efficient population. 

We define two behavioral metrics for assessing cooperation between the members of the successful dyads: anti-synchrony and load-sharing.


\textit{Anti-synchrony} ($C_{\text{AS}}\in[0,1]$) is defined as the mean difference between the role of the two agents (Eq \ref{eq:metricAS}). This tells us the fraction of time both roles are filled (the agents assume complementary roles). Dyads that swap roles in a temporally coordinated way have $C_{\text{AS}}$ near 1.

\begin{equation}
    \label{eq:metricAS}
    C_{\text{AS}}(R_1, R_2) = \frac{1}{N}\Sigma_n |R_1 - R_2| \\
\end{equation}

\textit{Load-sharing} ($C_{\text{LS}}\in[0,1]$) is defined as the minimum ratio of the individual efforts (Eq \ref{eq:metricLS}) where $E_i$ is the RMS of the force signal. 

\begin{equation}
    \label{eq:metricLS}
    C_{\text{LS}}(E_1, E_2) = \min(\frac{E_1}{E_2}, \frac{E_2}{E_1}) \\
\end{equation}




Figure \ref{fig:pareto-scatter} illustrates how anti-synchrony and load-sharing emerge in the Pareto front (32772 dyads). We observe that the dyads with the best performance (lowest tracking and stabilization loss) had high load-sharing and high anti-synchrony. Load-sharing evolved ubiquitously, however, found in both well and poorly tracking dyads; 90\% of dyads exhibit load-sharing above 0.75. This property is likely an artifact of the task and selection of controllers, an unintended (or unavoidable) consequence of these kinds of cooperative tasks. 

However, we observe the full spectrum of synchrony and anti-synchrony. And anti-synchrony is correlated to success in the task. While anti-synchrony is not sufficient for good performance, the dyads that perform best at \emph{both} tracking and stabilization are highly anti-synchronous (\fref{pareto-scatter}e-f). It is possible for dyads to track well with low anti-synchrony as is evinced by the vertical cluster at the left of \fref{pareto-scatter}d, these dyads routinely violate the force constraints (\fref{traces*}b). 

\begin{figure*}[tbh]
 \centering
 \includegraphics{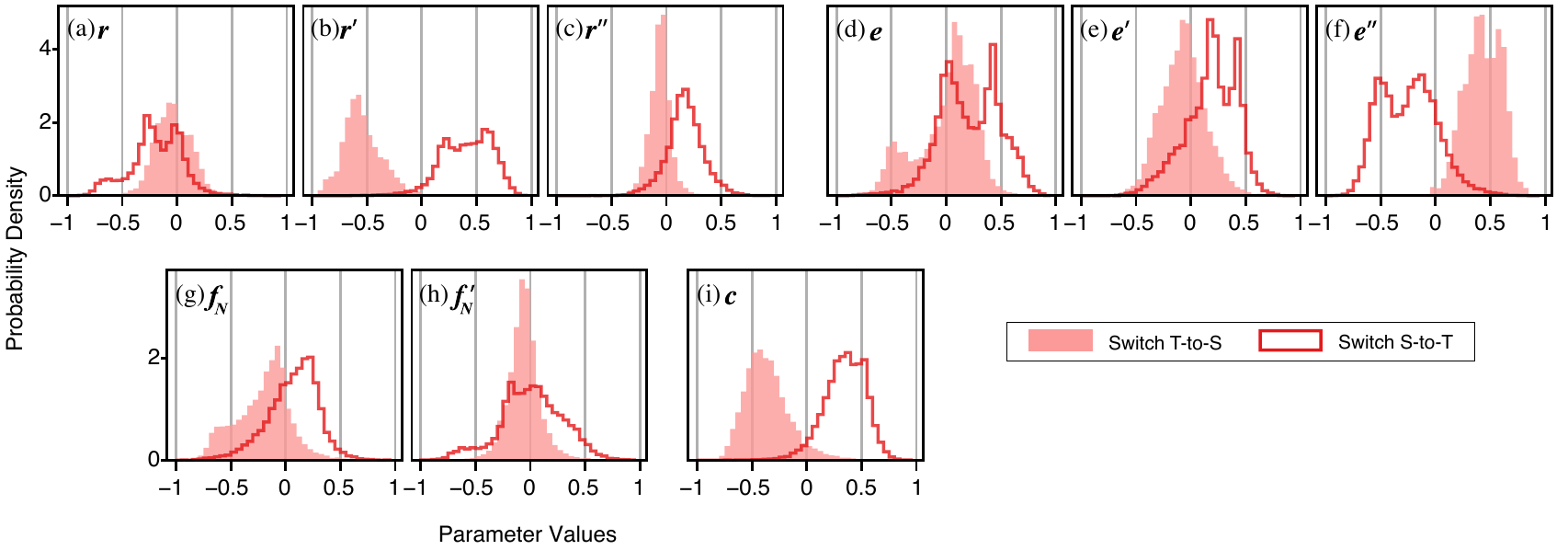}
 \caption{Parameter distributions of both S-to-T and T-to-S switching policies. \add{Each subfigure depicts the weight of a feature, labeled in bold, in the switching policies. Bimodality of feature weights (a,d,e,f) suggests multiple converged solutions within the Pareto-efficient population. Unimodal distributions with non-zero means indicate that the population largely converged on a single value for the given parameter. Zero-mean distributions suggest that the particular feature was not salient for that switching policy.}}
 \label{fig:dist*}
\end{figure*}

\subsection{Evolved symmetry}
To explore the parameter space of the Pareto-optimal population, we inspect the parameter distributions of the top-performing dyads (tracking loss $<0.7$, stabilization loss $<0.3$). The resulting set has 7359 dyads. Each dyad consists of 2 agents, each of which has a policy for switching from stabilization role to tracking role and vice versa. We aggregate the policy parameter vectors into two groups, T-to-S and S-to-T (each dyad contributes two of each, one from each agent). The histograms of each weight for both policies are summarized in \fref{dist*}. The weights for many features have diffuse distributions, indicating that there is significant variability in the switching policies. Some parameters have a clear non-zero mean, indicating that they play a significant role in many policies. In addition, some parameters have bimodal distributions (\fref{dist*}a, d, e, \& f), suggesting multiple distinct strategies. 

We define a \textit{symmetry} coefficient as a shifted version of angular similarity between two vectors. If  two policy vectors were identical, their symmetry would be 1 and conversely, if one were the negative of the other, their symmetry would be -1.
\begin{equation}
    \label{eq:metricSym}
    C_{\text{Sym}}(W_1, W_2) = 1-\frac{2}{\pi}\cos^{-1}\Big(\frac{\langle W_1,W_2\rangle}{|W_1|\cdot|W_2|}\Big)
\end{equation}

We apply this symmetry metric to dyads, comparing the similarity between the agents' policies, in four select sub-groups of the Pareto front: the dyads in the top and bottom decile of anti-synchrony  (\fref{symm*}a-c, \emph{green} and \emph{yellow}), the dyads in the top decile of load-sharing (d-f, \emph{red}), and an equally sized (10\% of the total population), randomly sampled population (a-d, \emph{black outline}). We compare the corresponding T-to-S and S-to-T policies between the agents, as well as a cross-comparison of one agent's $W_{TS}$ policy to the other's $W_{ST}$ (the average of both combinations). 

The most anti-synchronous dyads (\emph{green}) exhibit higher than the average symmetry comparing $W_{ST}$ policies between agents and anti-symmetry (more negative) in the cross comparison $C_{Sym}(W_{ST,i},W_{TS,j})$ between agents. In contrast, the most equitably load-sharing dyads are practically indistinguishable from the random sample with regards to symmetry (\fref{symm*}d-f).

\section{DISCUSSION AND CONCLUSION}

Our evolution generates a large variety of dyads that successfully solve a simulated cooperative physical task. This synthetic population reveals several interesting relationships between performance and attributes often associated with cooperation.

\subsection{Is anti-synchrony necessary for high performance?}
Certainly, anti-synchrony is not sufficient, as evidenced by the numerous dyads that exhibit high anti-synchrony and poor tracking or stabilization performance (\fref{pareto-scatter}d-f). However, of those dyads that performed best in stabilization (or stabilization and tracking), all exhibited high anti-syncrony. At the very least, the data are consistent with the claim that anti-synchrony might be a prerequisite for successful cooperation \emph{in this task}. Granted, by design, the reference trajectories and force constraints require both roles to be filled for the majority of the trial. Were the force boundaries relaxed, agents could co-exist in reference-tracking role achieving high performance (having no concern for the force-stabilization penalty) with no temporal coordination. Our task and the hypothetical counter example perhaps present two extremes on the demands of temporal coordination. How much anti-synchrony is required for cooperation is likely a function of many variables, including the task dynamics and the controllers of the cooperating agents. 
Future explorations
can characterize the task-specific determinants of anti-synchrony.

\subsection{Why symmetry?}
If one were tasked to invent cooperative dyads, symmetry would be an intuitive design paradigm. Composing pairings of identical agents with anti-symmetric policies for T-to-S and S-to-T transitions would lead to cooperative behaviors exhibiting perfectly equitable load sharing and precise anti-synchrony. As such, we presumed that load-sharing and anti-synchrony would somehow be the indicators of cooperation. Our intuition was wrong; neither attribute is sufficient for cooperation. While we did observe both attributes in high performing dyads, load sharing was not selective at all to that group. Equitable load sharing emerged ubiquitously across dyads of all levels of performance. In contrast, anti-synchrony was not expressed quite as universally. And interestingly, Figure \ref{fig:symm*}c confirms that the top decile of anti-synchronous dyads from our evolved populations exhibit that naively predicted pattern of symmetry and anti-symmetry. This suggests that our evolutionary algorithm may have arrived at the same intuitive genomic pattern giving rise to temporal coordination. However, it seems that many paths in parameter space arrive at load sharing.

\subsection{Sparse policies, policy clusters}
The histograms of parameter distribution (\fref{dist*}) show high entropy for most of the feature weights. Having a weight of magnitude 1 for a feature indicates that the other weights in the policy must be 0, since we normalize policy vectors to 1. We observe in the histogram weights near -1 in \fref{dist*}b (T-to-S) and \fref{dist*}f (S-to-T), suggesting the existence of sparse policies. Additionally, bimodality observed in the distribution of some feature weights suggests the existence of multiple unique policies that combine features differently. Systematic clustering in the parameter space may reveal distinct classes of switching policies. These classes represent testable hypotheses as to how humans may use visual and haptic information in cooperative tasks.

\subsection{Future validation in human subjects}
\add{Experiments with human subjects or mixed human-robot dyads will address whether the same patterns of anti-synchrony and symmetry emerge. In the above simulations, we use parsimonious models for the task dynamics and switching policies. Data from human dyadic cooperation will inform models that more faithfully represent human actors (\eg empirically fit tracking controllers, adaptation and learning, etc.).}

\bibliographystyle{IEEEtran.bst} 
\bibliography{refs_coop}

\end{document}